\documentclass[letterpaper, 10 pt, journal, twoside]{IEEEtran}
\usepackage{cite}
\usepackage{amsmath,amssymb,amsfonts}
\usepackage{algorithmic}
\usepackage{graphicx}
\usepackage{textcomp}
\usepackage{tabularx}  
\usepackage{ragged2e}  
\usepackage{makecell}  
\usepackage{array} 
\usepackage{mdframed}
\usepackage{comment}
\usepackage{url}
\usepackage[T1]{fontenc}
\usepackage{csquotes}
\usepackage{float} 
\usepackage{tabularx}  
\usepackage{booktabs}  
\usepackage{multirow}  
\usepackage{hyperref}
\usepackage{caption}
\usepackage[table]{xcolor} 

\newcolumntype{L}{>{\RaggedRight\arraybackslash}X} 

\def\BibTeX{{\rm B\kern-.05em{\sc i\kern-.025em b}\kern-.08em
    T\kern-.1667em\lower.7ex\hbox{E}\kern-.125emX}}

\IEEEoverridecommandlockouts
\title{RL-Driven Data Generation \\ for Robust Vision-Based Dexterous Grasping}
\author{
Atsushi Kanehira$^{1*}$,
Naoki Wake$^{1*}$,
Kazuhiro Sasabuchi$^{1}$,
Jun Takamatsu$^{1}$,
and Katsushi Ikeuchi$^{1}$
\thanks{$^{1}$Applied Robotics Research, Microsoft, Redmond, WA 98052, USA
        {\tt\footnotesize atsushi.kanehira@microsoft.com}}%
\thanks{*Equal contribution.}
\thanks{Digital Object Identifier (DOI): see top of this page.}
}

\begin{document}
\maketitle
\begin{abstract}

This work presents reinforcement learning (RL)-driven data augmentation to improve the generalization of vision-action (VA) models for dexterous grasping. While real-to-sim-to-real frameworks—where a few real demonstrations seed large-scale simulated data—have proven effective for VA models, applying them to dexterous settings remains challenging: obtaining stable multi-finger contacts is nontrivial across diverse object shapes. To address this, we leverage RL to generate contact-rich grasping data across varied geometries. In line with the real-to-sim-to-real paradigm, the grasp skill is formulated as a parameterized and tunable reference trajectory refined by a residual policy learned via RL. This modular design enables trajectory-level control that is both consistent with real demonstrations and adaptable to diverse object geometries. A vision-conditioned policy trained on simulation-augmented data demonstrates strong generalization to unseen objects, highlighting the potential of our approach to alleviate the data bottleneck in training VA models.

\end{abstract}


\section{Introduction}
\IEEEPARstart{R}{ecent} progress in vision-action (VA) models has enabled flexible dexterous manipulation, driven by expressive architectures (e.g., diffusion models) and large-scale demonstration datasets~\cite{brohan2023rt2, zhu2023vima, chi2023rss, patel2023corl}. Yet, VA models still struggle to generalize to unseen object shapes and task settings without fine-tuning on additional real-world data~\cite{xie2023decomposing, lee2022vrb, fu2023robotdex}. This issue is acute in dexterous manipulation, where collecting expert demonstrations is costly and difficult to scale~\cite{mandlekar2023mimicgen, janner2022neurips}.

This work investigates how simulation can help alleviate the data bottleneck in dexterous imitation learning by generating diverse grasping trajectories across a wide range of object shapes. Our aim is to expand training data in a scalable and geometry-aware manner, enabling robust vision-based policies without relying on extensive real-world demonstrations.

As a foundation for our study, we draw on prior research that has explored simulation-based augmentation of demonstrations using small numbers of human-provided examples~\cite{mandlekar2023corl, liu2024icra}. These methods typically segment demonstrations into sub-skills and synthesize new trajectories by varying object poses and scene configurations. While some studies have extended this to dexterous manipulation scenarios~\cite{liu2024icra}, finger motions are reused directly from the original demonstrations, implicitly assuming that object shape remains constant.
Given that achieving stable grasps in dexterous manipulation requires carefully coordinated finger placements that are highly sensitive to object shape~\cite{rajeswaran2018learning, andrychowicz2020learning, chen2022system}, this assumption imposes a strong limitation on the diversity of data that can be generated in simulation.

To overcome this limitation, we explore an approach that uses reinforcement learning (RL) to autonomously generate grasping behaviors adapted to different object shapes. The RL policy is trained in simulation using privileged information such as object pose and contact states~\cite{hu2024iclr}, enabling it to acquire flexible finger motions that respond to geometric variation. This allows for the generation of diverse and realistic training data that goes beyond the limitations of demonstration-based replay~\cite{li2023iclr, ha2023arxiv}.

\begin{figure}[t]
  \centering
  \includegraphics[width=0.49\textwidth]{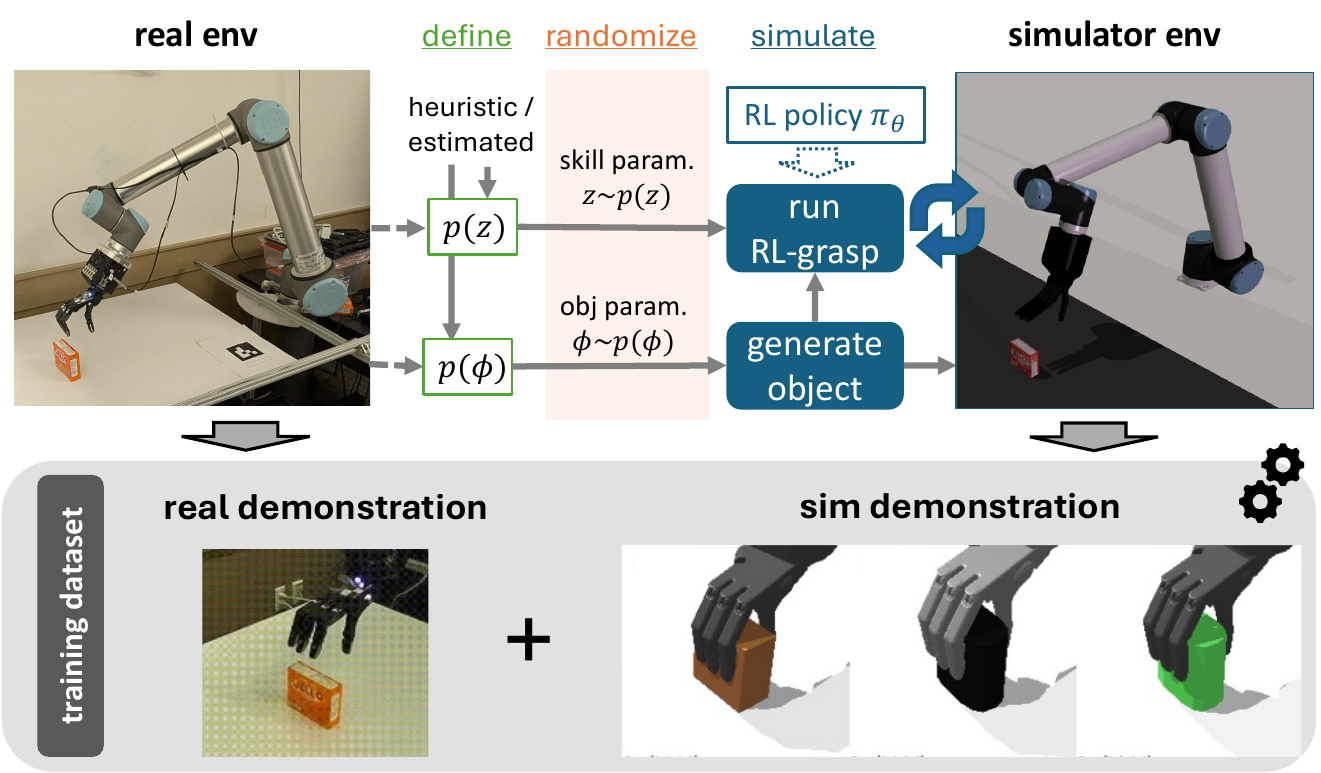}
\caption{
  \small
  Overview of the RL-driven data collection framework. Grasping trajectories are generated by executing a parameterized grasp skill in simulation. The skill uses a residual RL policy trained with privileged information and is conditioned on a skill parameter \( z \) (e.g., approach direction). The simulated environment includes a single object, parameterized by \( \phi \). Both \( z \) and \( \phi \), along with other simulation factors (e.g., camera pose and lighting), are sampled from distributions derived from heuristics or estimated from real demonstrations. The resulting simulated trajectories are used to train a vision-to-action imitation learning model.
  }
  \label{fig:intro}
  \vspace{-1.5em}
\end{figure}

Our approach is based on a hybrid execution framework similar to~\cite{lee2023skillgen}, which decomposes execution into two phases: a \textit{non-contact phase}, handled by conventional motion planning (e.g., RRT) and a \textit{contact phase}, governed by a learned grasp skill involving hand-object interactions. The grasp skill operates by combining a reference trajectory with a residual policy learned through RL. The reference trajectory is determined by skill parameters, such as approach direction, and specifies a nominal motion for the end-effector and finger joints. The residual RL policy then outputs displacements that adapt the motion to the current object instance.
This separation provides two key benefits: (1) it simplifies and stabilizes RL training by reducing search space, and (2) it enables trajectory-level modulation via skill parameters, making the overall behavior easier to align with human-provided demonstrations and, thereby, open to real-to-sim-to-real frameworks. The shape of target objects is parameterized using superquadrics, and both object parameters and skill parameters are sampled from distributions designed to reflect the realistic variations encountered in the physical world. This sampling strategy promotes behavioral diversity while ensuring that the generated trajectories remain plausible for real-world execution.

We further conduct experiments in a real-world robotic environment using a controlled dataset, allowing us to evaluate generalization performance to unseen object shapes and textures. Our results demonstrate that mixing the simulated data generated by this pipeline with a small amount of real-world data significantly improves grasping robustness to both visual and geometric variations.

\section{Related Work}

\subsection{Dexterous Grasping with Vision-Based Policies}

Dexterous manipulation with multi-fingered hands presents unique challenges in perception and control~\cite{kang1997toward}. Vision-based policies address this by learning expressive mappings from visual inputs to high-dimensional motor commands. Diffusion Policy~\cite{chi2023rss} introduced a denoising diffusion framework for imitation learning, enabling multi-modal action prediction with stable training. Building on this, GraspDiff~\cite{patel2023corl} applies diffusion to generate diverse grasp configurations conditioned on vision. DexMV~\cite{janner2022neurips} learns dexterous policies from multi-view human video, while MetaGrasp~\cite{gupta2022corl} uses meta-learning to adapt to novel objects with few examples. DexGraspNet~\cite{sharma2023iros} provides a large-scale synthetic dataset for benchmarking dexterous grasping. These methods often rely on simulation because of the high cost of real-world data collection. General-purpose architectures such as CLIPort~\cite{shridhar2023ral} and VIMA~\cite{zhu2023iclr} have also been applied to dexterous control, although training remains sample-inefficient.

\subsection{Foundation Models for Generalist Robotic Control}

Foundation models aim to generalize across tasks and embodiments by leveraging massive, heterogeneous data. RT-1 and RT-2~\cite{brohan2023rt2} train vision-language-action policies on over 130k real-world demonstrations. RoboCat~\cite{reed2022arxiv} and PaLM-E~\cite{driess2023ral} combine vision, language, and proprioception for embodied reasoning. Open X-Embodiment~\cite{openx2023arxiv} and Octo~\cite{openx2023octo} scale pretraining to over a million demonstrations across diverse robot types. GR00T~\cite{nvidia2024arxiv} further explores multi-timescale planning for long-horizon tasks. While these models exhibit broad generalization, data collection remains a bottleneck—especially for dexterous manipulation. Several studies~\cite{xie2023corl, lee2022vrb, fu2023robotdex} highlight the vulnerability to changes in viewpoint, lighting, or context, motivating the need for robust and data-efficient training strategies.

\subsection{Simulation-Based Data Generation and Augmentation}

Simulation has become essential for scaling data collection in dexterous learning. MimicGen~\cite{mandlekar2023corl} augments human demonstrations by segmenting them into sub-skills and synthesizing new trajectories via SE(3) variation. DexMimicGen~\cite{liu2024icra} extends this to bimanual tasks, generating over 20,000 demonstrations with high real-world success. SkillGen~\cite{lee2023ral} composes learned primitives via planning, while IntervenGen~\cite{jain2023corl} refines trajectories through corrective feedback. These systems reduce reliance on real-world teleoperation and enable large-scale data diversity.

In addition, simulator-privileged signals (e.g., object pose, contact forces) are leveraged to train expert policies, which are then distilled into vision-based models~\cite{hu2024iclr, li2023iclr}. Techniques like Eureka~\cite{huang2023neurips} automate reward design, while language-guided generation~\cite{ha2023arxiv} enables more flexible skill specification. Complementary to behavioral data, simulation also supports robust visual augmentation. GenAug~\cite{yang2023arxiv}, VRB~\cite{lee2022vrb}, and BridgeData V2~\cite{ebert2023rss} leverage domain and task variation to improve generalization. SE(3) diffusion~\cite{ni2023corl} and grasp diffusion~\cite{patel2023corl} enable physically plausible perturbations in object and hand pose. Collectively, these methods enable scalable and generalizable learning of dexterous skills with minimal reliance on costly real-world data.

\section{RL-driven Data Generation Framework}

To improve the generalization and robustness of vision-based policies for dexterous grasping, we aim to augment training data through simulation. In particular, we generate diverse and physically plausible grasping trajectories using a RL-driven grasp skill, and use the resulting trajectories as demonstrations for downstream imitation learning.

The data collection process is structured as a two-stage grasping pipeline. The \textit{non-contact phase}—in which the robot moves toward the object—is executed using conventional motion planning. The \textit{contact phase}, which involves direct interaction with the object and grasp execution, is governed by a parameterized grasp skill. This decomposition allows us to focus learning capacity on the complex, contact-driven portion of the motion, while preserving deterministic planning for the approach phase. In the remainder of this section, we focus exclusively on the second stage—the contact-rich phase—as it is the source of data used for training vision-based policies.

\subsection{Problem Formulation}
\label{sec:problem_formulation}

We aim to collect a dataset \( \mathcal{D} = \{\tau_i\}_{i=1}^N \), where each trajectory \( \tau = \{(\hat{s}_t, \hat{a}_t)\}_{t=0}^T \) consists of a sequence of robot-observable state-action pairs. Here, \( \hat{s}_t \in \hat{\mathcal{S}} \) denotes the observable state (e.g., RGB images and proprioception), and \( \hat{a}_t \in \hat{\mathcal{A}} \) is the corresponding control command (e.g., end-effector poses or finger joint deltas).

These trajectories are generated by executing a parameterized grasp skill function \( f_{\text{grasp}} \), which maps a simulator state \( s_t \), a skill parameter \( z \), and an object instance \( o \) to a low-level control action \( a_t \) (defined in Section~\ref{sec:grasp-skill}).

We assume the environment contains a single target object $o$, whose shape is parameterized by a compact descriptor \( \phi \in \Phi \). The specific parameterization of object shape is described in Section~\ref{sec:object-shape}.

Each simulator action \( a_t \) is projected into an observable action \( \hat{a}_t \), and paired with the corresponding \( \hat{s}_t \), yielding a training example for vision-based imitation learning. This transformation is straightforward and serves to convert simulator-internal representations into forms compatible with real-world robot sensors and actuators.

\subsection{Grasp Skill Parameterization}
\label{sec:grasp-skill}
\begin{figure}[!t]
  \centering
  \includegraphics[width=0.48\textwidth]{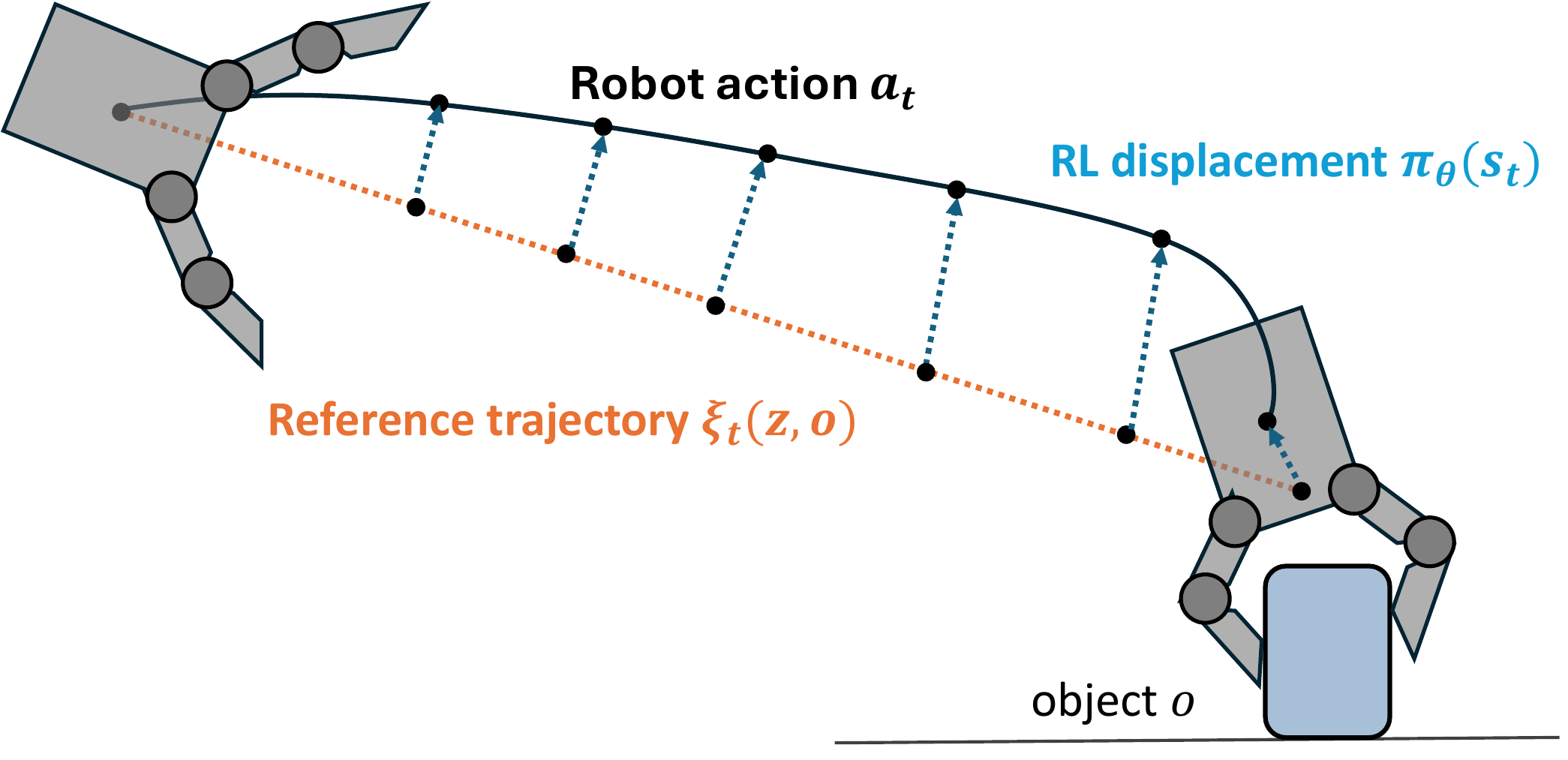}
  \vspace{-0.5em}
  \caption{\small
Schematic illustration of grasp skill execution. The reference trajectory \( \xi_t(z, o) \) (orange dots), which defines the nominal end-effector pose and finger joint configuration, is refined at each timestep by the residual policy \( \pi_\theta(s_t) \) (blue dots). The final robot action \( a_t = \xi_t(z, o) + \pi_\theta(s_t) \) (black) incorporates both position and finger corrections. }

  \label{fig:rl}
\vspace{-1.5em}
\end{figure}

The grasp skill is defined as a function \( f_{\text{grasp}}(s_t, z, o) \) that outputs a low-level control action \( a_t \) by combining a precomputed reference trajectory and a residual correction policy trained via RL. As illustrated in Fig.~\ref{fig:rl}, the action at each time step \( t \) is given by:
\[
a_t = f_{\text{grasp}}(s_t, z, o) = \xi_t(z, o) + \pi_\theta(s_t),
\]
where:
\begin{itemize}
  \item \( \xi_t(z, o) \) is a reference trajectory defined by a skill parameter \( z \) (e.g., approach direction) and the current object \( o \),
  \item \( \pi_\theta(s_t) \) is a residual policy that produces corrective actions based on the simulator state \( s_t \),
  \item \( \theta \) denotes the parameters of the neural network trained via RL.
\end{itemize}

This two-part structure has two key benefits. First, the reference trajectory encodes human prior knowledge, improving the sample efficiency and stability of RL. Second, the use of an explicit, externally controllable skill parameter \( z \) enables structured variation in the generated trajectories. This makes it possible to guide data generation toward distributions that better match real-world conditions, which is particularly useful for simulation-based augmentation.

\subsection{Object Shape Representation}
\label{sec:object-shape}

To generate grasping trajectories for a diverse range of objects, we represent object geometry using superquadric shapes---a compact and differentiable parametric family that can model a wide variety of convex and concave forms.

A superquadric is defined implicitly by the equation:
\[
F(x, y, z; \phi) = \left( \left| \frac{x}{a_1} \right|^{\frac{2}{\epsilon_2}} + \left| \frac{y}{a_2} \right|^{\frac{2}{\epsilon_2}} \right)^{\frac{\epsilon_2}{\epsilon_1}} + \left| \frac{z}{a_3} \right|^{\frac{2}{\epsilon_1}} - 1 = 0,
\]
where the shape parameters \( \phi = \{a_1, a_2, a_3, \epsilon_1, \epsilon_2\} \) define the object's scale along each axis and the degree of curvature or squareness in each direction.

\begin{figure}[t]
  \centering
  \includegraphics[width=0.48\textwidth]{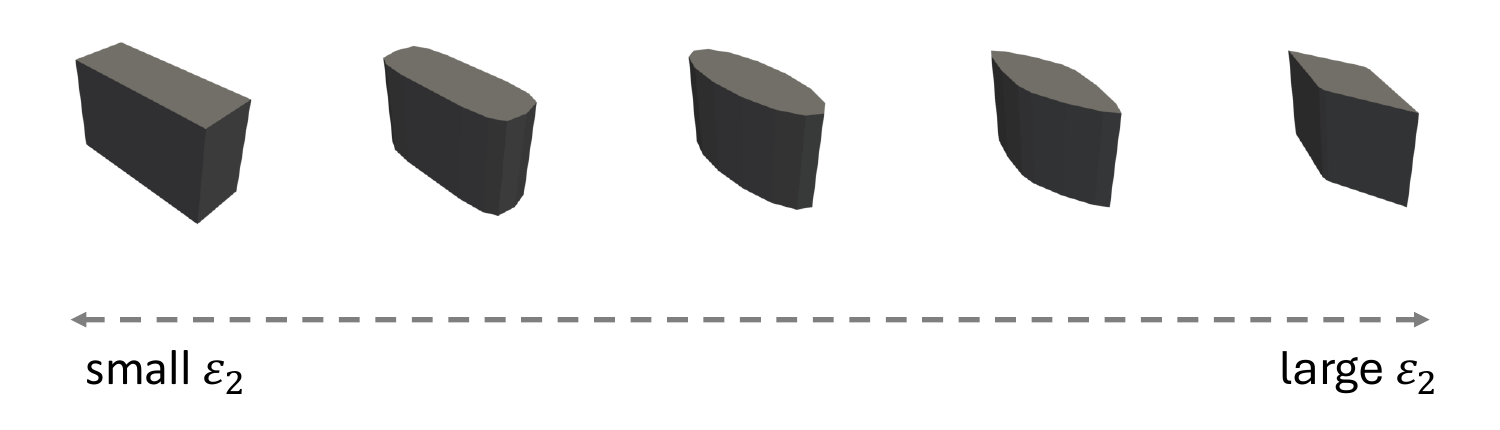}
  \vspace{-0.5em}
  \caption{\small
Visualization of object shape variation by sweeping the superquadric parameter \( \varepsilon_2 \) from a small positive value to 2.0. Each shape is rendered with the same scale, and all other parameters are fixed. This illustrates how geometric smoothness and edge sharpness change as \( \varepsilon_2 \) increases.
}
  \label{fig:sq}
\vspace{-1.5em}
\end{figure}

\begin{itemize}
    \item The scale parameters \( a_1, a_2, a_3 \in \mathbb{R}^+ \) control the size of the object along the \( x, y, z \) axes.
    \item The shape exponents \( \varepsilon_1, \varepsilon_2 \in (0, 2] \) control the roundness of the object: lower values produce pinched shapes, while higher values yield boxy forms (see Fig.~\ref{fig:sq}).

\end{itemize}
This representation allows us to smoothly interpolate across different shapes in a low-dimensional parameter space, making it well-suited for generating diverse yet structured grasping scenarios.

\subsection{Data Collection and Parameter Sampling}
\label{sec:data-collection}

To construct the imitation learning dataset, we simulate grasping episodes by randomly sampling:

\begin{itemize}
  \item a skill parameter \( z \sim p(z) \), which specifies the intended grasp configuration (e.g., approach direction),
  \item an object shape parameter \( \phi \sim p(\phi) \), which defines the geometry of the single object present in the scene.
\end{itemize}

At each time step \( t \), the simulator computes the control action using the learned grasp skill \(a_t = f_{\text{grasp}}(s_t, z, o)\). The corresponding robot-observable state \( \hat{s}_t \) includes sensory inputs such as RGB images and proprioceptive signals, while the projected action \( \hat{a}_t \) comprises control commands such as end-effector poses and finger joint displacements. Each trajectory \( \tau = \{(\hat{s}_t, \hat{a}_t)\}_{t=0}^T \) is added to the dataset \( \mathcal{D} \) only if a task-specific success condition (e.g., lifting without slippage) is met. This procedure enables efficient generation of diverse and physically plausible grasp demonstrations that match the input-output structure of vision-based imitation learning policies.

The distributions \( p(z) \) and \( p(\phi) \) can be estimated or determined heuristically. For example, the skill parameter \( z \) can be estimated from teleoperation demonstrations, or extracted from human execution videos using a skill-specific encoder within a Learning-from-Observation (LfO) framework. The object parameter \( \phi \) is derived from prior knowledge of the environment or could be estimated using the predictor (e.g., SuperDec~\cite{fedele2025superdec}), which decomposes 3D scenes into superquadric primitives from pointcloud observations.

These estimations allow our simulated data generation pipeline to be aligned with realistic human and object distributions, making it compatible with real-to-sim-to-real learning paradigms.

\section{Experiment}

\begin{figure}[!t]
  \centering
  \includegraphics[width=0.48\textwidth]{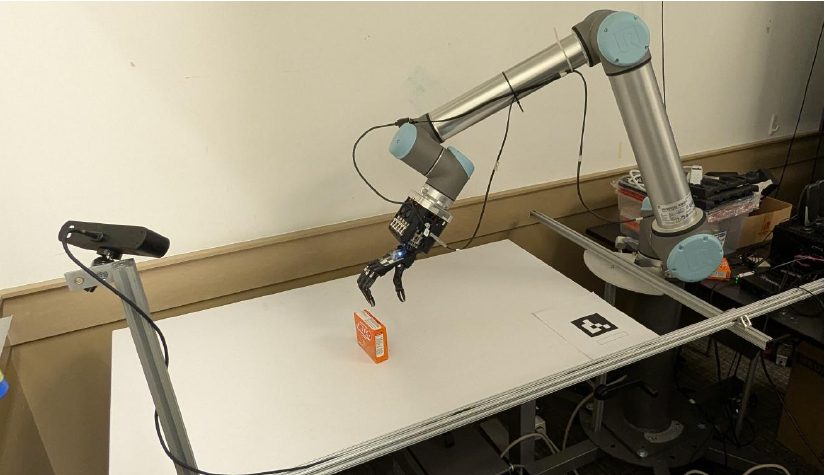}
  \vspace{-0.2em}
  \caption{\small
  Real-world setup used for evaluation. A UR10e arm with a Shadow Dexterous Hand Lite performs grasping tasks while receiving RGB input from a fixed ZED2i camera.}
  \label{fig:env}
    \vspace{-1.5em}
\end{figure}

We conducted real-world experiments to evaluate whether simulation-augmented data improve generalization in vision-based dexterous grasping. We first describe the hardware setup and task specification in Section~\ref{sec:hardware-task}. Then, we explain the details of our simulation-based data collection process in Section~\ref{sec:data-collection}, followed by the imitation learning setup in Section~\ref{sec:imitation-learning}. Finally, Section~\ref{sec:real-world-eval} presents our experimental results and discussion.

\subsection{Environment Setup}\label{sec:hardware-task}
\noindent\textbf{Hardware Setup:}
Our physical setup consists of a 6-DoF UR10e robotic arm (Universal Robots) equipped with a Shadow Dexterous Hand Lite (Shadow Robotics), a four-fingered anthropomorphic hand as shown in Fig.~\ref{fig:env}. While the hand lacks built-in tactile sensors, torque readings from one joint per finger (FF3, MF3, RF3, and TH2, as defined in the manufacturer's documentation\footnote{\url{https://www.shadowrobot.com/dexterous-hand/}}) were used as proprioceptive feedback.
A ZED2i stereo camera (StereoLabs) was mounted in a fixed position relative to the robot to provide visual input. Although the ZED2i supports depth sensing, we used only the left RGB image stream to match the input modality expected by the pretrained VA model used in this study.

\vspace{0.5em}
\noindent\textbf{Task Setup:}
The robot was tasked with performing top-down grasps on tabletop objects. A trial was considered successful if the object was grasped and lifted without slippage. Each object was evaluated over five independent trials.

\subsection{Data Collection Setup}\label{sec:data-collection}
\noindent\textbf{Reference Trajectory Implementation.}
The reference trajectory \( \xi\) is constructed by interpolating between a pre-grasp pose and a grasp pose, both of which include the end-effector position and orientation as well as finger joint configurations.

We define a full pose as a tuple \( x = (R, \mathbf{p}, \mathbf{q}) \), where \( R \in SO(3) \) is the end-effector orientation, \( \mathbf{p} \in \mathbb{R}^3 \) is the position, and \( \mathbf{q} \in \mathbb{R}^n \) represents the finger joint states. The grasp pose \( x_g = (R_g, \mathbf{p}_g, \mathbf{q}_g) \) is computed based on the object geometry \( o \), while the pre-grasp pose \( x_0 = (R_0, \mathbf{p}_0, \mathbf{q}_0) \) is determined by applying an approach direction specified in the skill parameter \( z \).

The reference trajectory at time step \( t \) is then computed by interpolating between the two poses:
\[
\xi_t(z, o) = \texttt{interpolate}(x_0, x_g; t/T),
\]
where \( T \) denotes the predefined total number of timesteps for the trajectory. Grasp pose is estimated via desired finger-tip pose from ContactWeb~\cite{kang1997toward}.

This produces a smooth nominal trajectory that moves the hand from the pre-grasp pose to the grasp pose over time. While more complex interpolation schemes could be used, we adopt this simple approach for ease of implementation and generality across different object instances. When the object pose is randomized via SE(3) transformations (as in MimicGen~\cite{mandlekar2023mimicgen} or DexMimicGen~\cite{liu2024dexmimicgen}), the reference trajectory is transformed accordingly.

\begin{figure}[!t]
  \centering
  \includegraphics[width=0.45\textwidth]{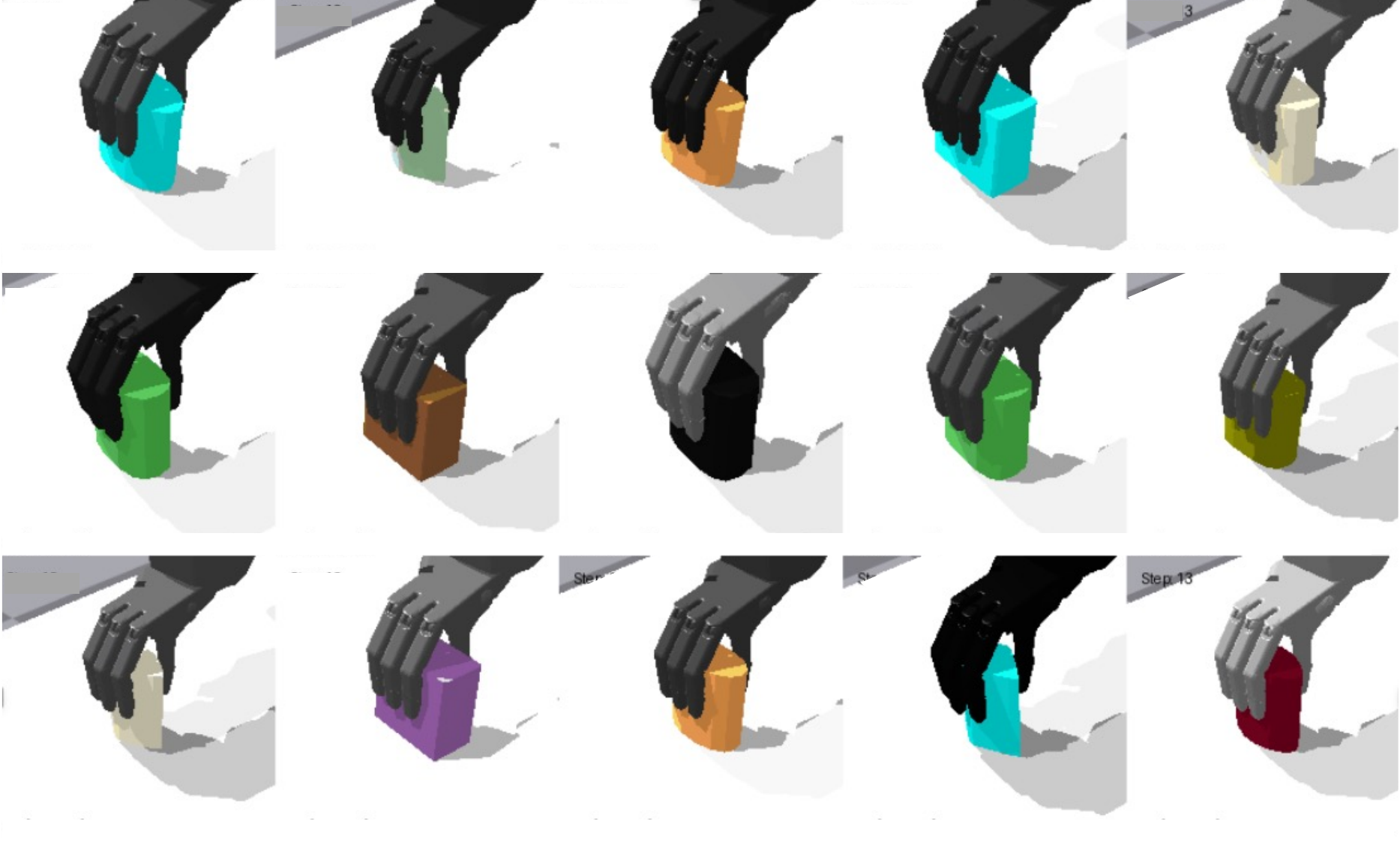}
  \vspace{-1em}
\caption{\small
Example simulated grasp outcomes generated by our framework across diverse object shapes, used as training data for vision-based policies.
}
  \label{fig:data}
\vspace{-1.5em}
\end{figure}

\vspace{0.5em}
\noindent\textbf{RL-Based Residual Policy Training:}
We train the residual grasp policy \( \pi_\theta \) using Proximal Policy Optimization (PPO)~\cite{schulman2017ppo} within the IsaacGym simulator~\cite{makoviychuk2021isaac} following the setting of~\cite{10000167}. The policy operates on simulator-observable states and outputs residual actions that modulate a pre-defined reference trajectory.

The state is a low-dimensional vector that encodes the system's physical configuration: hand pose (in world and hand frames), object pose relative to the hand, fingertip positions in multiple coordinate frames, joint angles, contact indicators from proximity and force signals, and fingertip force directions. The action specifies residual displacements to the end-effector pose and finger joint angles relative to a reference trajectory. The reward function is composed of three terms:
\[
r = r_{\text{dist}} + r_{\text{force}} + r_{\text{pick}},
\]
where:
\begin{itemize}
  \item \( r_{\text{dist}} \) encourages each fingertip to approach the predicted contact points on the object surface.
  \item \( r_{\text{force}} \) rewards stable contact by promoting appropriate contact forces at the fingertips.
  \item \( r_{\text{pick}} \) provides a binary bonus when the object is successfully lifted above a specified height threshold.
\end{itemize}

\vspace{0.5em}
\noindent\textbf{Simulation Parameter Sampling:}
We define the sampling distributions \( p(z) \) and \( p(\phi) \) to generate diverse and realistic trajectories in simulation.

For the skill parameter \( p(z) \),
the grasp approach direction was estimated from real demonstration trajectories. Based on the observed distribution, we defined \( p(z) \) as a uniform distribution centered at 80° from table surface with \(\pm10^\circ\) variation.

For the object parameter \( p(\phi) \),
each object shape is represented by a superquadric parameter vector \( \phi = [a_1, a_2, a_3, \varepsilon_1, \varepsilon_2] \).
The scale parameters \( a_1 \) and \( a_2 \) were sampled uniformly between 2–5~cm, and \( a_3 \) between 5–10~cm.
The shape exponents \( \varepsilon_1 \) and \( \varepsilon_2 \) were sampled from a uniform distribution between \( \varepsilon_{\text{min}} \) and 2.0, where \( \varepsilon_{\text{min}} \) is a small positive constant used to avoid degenerate shapes.

In addition, we also randomized other environmental factors, including lighting conditions, camera pose, object pose, and surface textures, to promote visual diversity and enhance robustness in downstream vision-based policy learning. Fig.~\ref{fig:data} shows sample frames from the simulated grasping trajectories used to train the imitation learning policy.

\subsection{Imitation Learning Setup}\label{sec:imitation-learning}
We fine-tune the Octo-Medium model---a 93M parameter transformer-based policy pretrained on 800k robot trajectories from the Open X-Embodiment dataset~\cite{openx2023octo}---using our collected dataset comprising both real-world and simulated grasping trajectories.

The policy takes as input \( \hat{s}_t \), which consists of a cropped RGB image of size \( 400 \times 400 \) and binarized fingertip contact indicators. It outputs the action \( \hat{a}_t \), a low-dimensional control vector composed of a 6D end-effector displacement (position and rotation deltas) and the absolute joint positions of the fingers.
Fine-tuning is conducted over 20,000 iterations using a behavior cloning loss, with default hyperparameters from the Octo codebase. During this process, only the final transformer layers are updated, while the vision-language encoder remains frozen.

\begin{figure}[!t]
  \centering
  \includegraphics[width=0.48\textwidth]{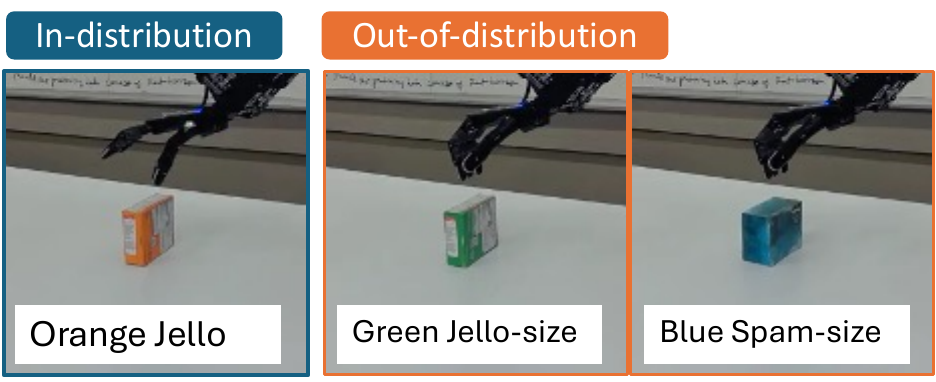}
  \caption{\small
Objects used for real-world evaluation. The \textbf{in-distribution} object (left) corresponds to the one used for collecting real-world demonstrations. The remaining \textbf{out-of-distribution} objects (right) differ in shape, size, or color and are used to test generalization. }

  \label{fig:objects}
  \vspace{-1.5em}
\end{figure}

\subsection{Real-World Evaluation}\label{sec:real-world-eval}

We designed a real-world evaluation to assess how simulated data affects grasping performance when mixed with limited real-world demonstrations. Specifically, we investigated the effectiveness of simulation-augmented training by varying the ratio of simulated and real trajectories during policy learning, and evaluated generalization to unseen objects.

\vspace{0.5em}
\noindent\textbf{Objects:}
We selected four physical objects for evaluation. One object—the orange jello-sized box—was also used during teleoperated data collection and thus represents an in-distribution (ID) case. The remaining objects differ in color, size, or shape and serve as out-of-distribution (OOD) test cases. These object types are shown in Fig.~\ref{fig:objects}.

\vspace{0.5em}
\noindent\textbf{Training Variants:}
We prepared three training conditions to isolate the contribution of simulated data:
\begin{enumerate}
  \item \textbf{Real-only:} 40 teleoperated demonstrations collected using the in-distribution object.
  \item \textbf{Sim-only:} 4000 simulated trajectories generated on simuilator environment.
  \item \textbf{Mixed:} A combination of 40 real and 4000 simulated demonstrations.
\end{enumerate}

Table~\ref{tab:real-world-results} summarizes the performance of policies trained with different data sources—\textbf{sim-only}, \textbf{real-only}, and \textbf{mixed}—evaluated separately on in-distribution and out-of-distribution objects.

The \textbf{sim-only} policy fails completely in real-world execution, achieving a 0\% success rate across all test objects. This clearly highlights the sim-to-real gap and indicates that policies trained solely in simulation do not transfer effectively without additional adaptation.

The \textbf{real-only} policy, trained on 40 demonstrations of a single object, achieves 100\% success on that object and also generalizes to one of the unseen objects, resulting in a 40\% success rate on out-of-distribution cases. While this demonstrates robustness within the training distribution, it also reveals limited generalization to new object shapes and sizes.

\begin{figure*}[!t]
  \centering
  \includegraphics[width=\textwidth]{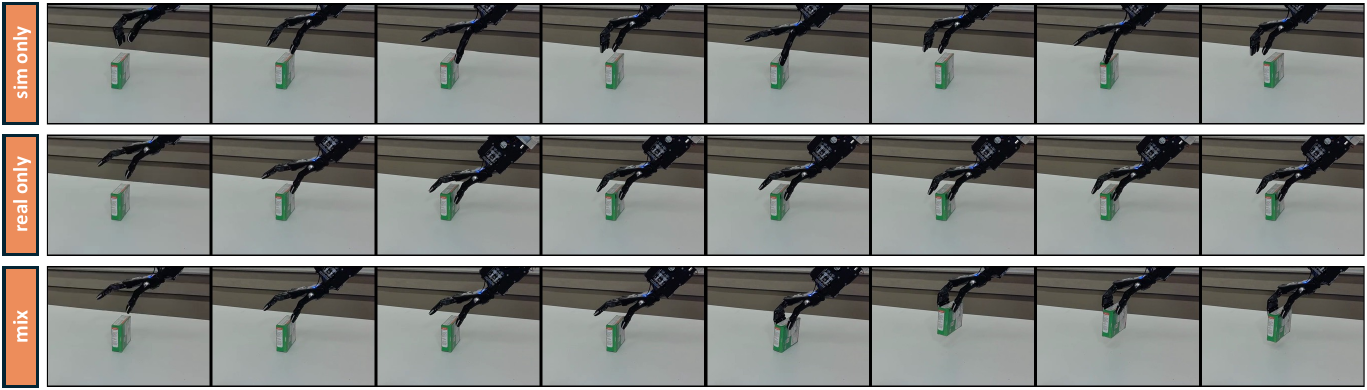}
   \vspace{-1.5em}
\caption{\small Qualitative comparison of real-world grasping results for different training settings. The \textbf{sim-only} policy fails to execute successful grasps, while the \textbf{real-only} policy performs well on seen objects but struggles to generalize. Only the \textbf{mixed} policy, trained with both simulated and real-world data, consistently achieves successful and robust grasps across diverse object shapes.}
  \label{fig:exp}
  \vspace{-1.5em}
\end{figure*}

In contrast, our proposed \textbf{mixed} policy—trained with both simulated and real-world data—achieves 100\% success across all tested objects, including those with unseen geometry, size, and texture. This result suggests that simulation-augmented data substantially improves generalization, even when only a small number of real-world demonstrations are available. Qualitative results shown in Fig.~\ref{fig:exp} further illustrate that only the mixed policy consistently succeeds in executing stable grasps, while the sim-only and real-only policies fail under out-of-distribution conditions. The diversity introduced by simulation, combined with grounding in real-world data, provides a scalable and effective strategy for dexterous grasp learning.
\begin{table}[!t]
\centering
\small
\begin{tabular}{lccc}
\toprule
\textbf{Object} & \textbf{real-only} & \textbf{ours (mixed)} & \textbf{sim-only} \\
\midrule
Orange jello-size box  & 5/5 (100\%) & 5/5 (100\%) & 0/5 (0\%) \\
Green jello-size box & 0/5 (0\%)   & \textbf{5/5 (100\%)} & 0/5 (0\%) \\
Blue spam-size box    & 1/5 (20\%)  & \textbf{5/5 (100\%)} & 0/5 (0\%) \\
\midrule
\textbf{Total (ID)}        & 5/5 (100\%) & 5/5 (100\%) & 0/5 (0\%) \\
\textbf{Total (OOD)}       & 1/10 (10\%) & \textbf{10/10 (100\%)} & 0/10 (0\%) \\
\bottomrule
\end{tabular}
\vspace{-0.5em}
\caption{\small
Real-world grasp success rates on in-distribution (ID) and out-of-distribution (OOD) objects. Each object was evaluated over 5 trials. The bottom rows report the total success rates for ID and OOD subsets respectively.
}
\vspace{-1.5em}
\label{tab:real-world-results}
\end{table}

\section{Conclusion}

We presented a simulation-driven framework for augmenting training data in vision-based dexterous grasping. Our approach combines skill-conditioned reference trajectories with a residual policy learned via RL, enabling the autonomous generation of diverse grasping behaviors in simulation. By mixing this simulated data with a small amount of real-world demonstrations, we trained a vision-based policy that generalizes effectively to unseen object shapes, sizes, and textures.

Real-world experiments demonstrated that policies trained with simulation-augmented data significantly outperform those trained with real or simulated data alone. Our results suggest that simulation is a powerful tool for scalable and robust policy learning in high-dimensional manipulation tasks.

\bibliographystyle{ieeetr}
\bibliography{bib}
\end{document}